\documentclass[runningheads]{llncs}

\usepackage{eccv}

\usepackage{times}
\usepackage{epsfig}
\usepackage{booktabs}
\usepackage{graphicx}
\usepackage{amsmath}
\usepackage{amssymb}
\usepackage{diagbox}
\usepackage{xcolor}
\usepackage{colortbl}
\definecolor{claude_red}{RGB}{255,0,0}
\newcommand{\ours}[1]{Un-EVIMO}
\usepackage{amsmath}
\DeclareMathOperator*{\argmax}{arg\,max}
\DeclareMathOperator*{\argmin}{arg\,min}
\usepackage{bbm}
\usepackage{arydshln}
\usepackage{xcolor,pifont}
\definecolor{mygreen}{RGB}{68, 117, 122}
\definecolor{myred}{RGB}{212, 76, 60}
\newcommand*\colourcheck[1]{%
  \expandafter\newcommand\csname #1check\endcsname{\textcolor{mygreen}{\ding{52}}}%
}
\colourcheck{green}
\newcommand*\colourno[1]{%
  \expandafter\newcommand\csname #1no\endcsname{\textcolor{myred}{\ding{55}}}%
}
\colourno{black}

\usepackage{eccvabbrv}

\usepackage{graphicx}
\usepackage{booktabs}

\usepackage[accsupp]{axessibility}  %

\usepackage[pagebackref,breaklinks,colorlinks,citecolor=eccvblue]{hyperref}

\usepackage{orcidlink}

\begin{document}

\title{\ours{}: Unsupervised Event-based Independent Motion Segmentation}

\titlerunning{\ours{}}

\author{Ziyun Wang\inst{1}\orcidlink{0000-0002-9803-7949} \and
Jinyuan Guo\inst{1}\orcidlink{0000-0002-4220-1148} \and
Kostas Daniilidis\inst{1, 2}\orcidlink{0000-0003-0498-0758}}

\authorrunning{Wang et al.}

\institute{University of Pennsylvania \and
Archimedes, Athena RC\\}

\newif\ifshowmainpaper %
\showmainpapertrue
\newif\ifshowsupplementary
\showsupplementarytrue

\ifshowmainpaper
\maketitle
\begin{abstract}
Event cameras are a novel type of biologically inspired vision sensor known for their high temporal resolution, high dynamic range, and low power consumption. Because of these properties, they are well-suited for processing fast motions that require rapid reactions. Event cameras have shown competitive performance in unsupervised optical flow estimation. However, performance in detecting independently moving objects (IMOs) is lacking behind, although event-based methods would be suited for this task based on their low latency and HDR properties. Previous approaches to event-based IMO segmentation heavily depended on labeled data. However, biological vision systems have developed the ability to avoid moving objects through daily tasks without using explicit labels. In this work, we propose the first event framework that generates IMO pseudo-labels using geometric constraints. 
Due to its unsupervised nature, our method can flexibly handle a non-predetermined arbitrary number of objects and is easily scalable to datasets where expensive IMO labels are not readily available. Our approach shows competitive performance on the EVIMO dataset compared with supervised methods, both quantitatively and qualitatively. See the project website for details: \url{https://ziyunclaudewang.github.io/un-evimo/}.
\end{abstract}
    
\section{Introduction}
\label{sec:intro}
Biological visual systems show remarkable performance in identifying independently moving objects when the viewer is undergoing self-motion. Basketball players can catch a ball flying at high speed while running across the court. Insects have neurons optimized for detecting independent motion to search for prey or avoid threats~\cite{nordstrom2006insect}. Cross-species studies have found that biological systems have neurons that specialize in detecting looming motion, a special case of independent motion~\cite{wu2023neural}.
Scientists have found that certain parts of the visual field are involved in subtracting out self-motion to help identify moving objects~\cite{pitzalis2013functional}. In cognitive science, the ability to model or segment independently moving objects has been extensively studied~\cite{raudies2013modeling, rushton2005moving, layton2016neural, royden2010detection}. Human drivers have the ability to identify moving pedestrians and avoid them even when the car is traveling at high speed. 
Another consideration is the speed of camera and depth sensors, which has become the bottleneck of autonomous vision~\cite{li2020towards}. High-accuracy depth sensors, e.g. LIDAR,  are able to map rigid scenes but have to apply semantic segmentation in order to detect Independently Motion Objects (IMOs).

\begin{table}[tb]
    \centering
    \caption{Feature comparisons. \ours{} does not simplify the geometry by following the complete motion field model; it does not require manual labeling of IMO objects; it trains a network that performs inference on scenes without extensive tuning; and it runs inference at real-time without heavy optimization.}
    \begin{tabular}{c c c c c}
    \toprule
         & \shortstack{Fast \\ (Real-Time)} & \shortstack{Scalable \\ (No IMO Labels)}  & \shortstack{Minimal Tuning } & Full Motion Models  \\
         \midrule
    EMSGC     & \blackno & \greencheck & \blackno  & \blackno \\
    EVIMO Network & \greencheck & \blackno & \greencheck & \blackno  \\
    SpikeMS & \greencheck & \blackno & \greencheck  & -\\
    ESMS & \blackno & \blackno & \greencheck  & -\\
    \ours{} & \greencheck & \greencheck & \greencheck & \greencheck \\
    \bottomrule
    \end{tabular}
    \label{tab:feature}
\end{table}
 The recent development of event-based cameras has brought hope to these issues. Event cameras are able to record the log change of brightness of individual pixels asynchronously. These low-latency cameras allow for continuous monitoring of motion patterns of the scene. In this work, inspired by biological vision systems, we use an event camera as a silicon ``eye'' and tackle the IMO segmentation problem given a stream of events. CNN-based approaches have shown success in dense segmentation tasks. In this work, we use neural networks as our predictor to take advantage of their generalizability. The bottleneck of event-based algorithms is the need for a tremendous amount of labeled training data. However, if we examine how species acquired the ability to handle IMOs, the labels do not need to come from annotated binary masks. Actually, many studies have shown that the motion field itself contains enough information to differentiate between self-motion and independent motion~\cite{nordstrom2006insect, wu2023neural}. An important question is: \textit{Can we learn motion segmentation with event cameras without manual labels by looking at the motion pattern in the scene?} In this work, we propose a novel framework for training IMO segmentation networks in an unlabeled dataset. \ours{} is the first event-based learning framework  for IMO detection without being trained with manual labels. 
We use a geometric self-labeling method to generate binary IMO pseudo-labels that supervise the IMO segmentation network. Our framework uses off-the-shelf optical flow prediction and input depth to fit 3D camera motion using RANSAC for excluding IMO as outliers. 
IMO flow field is obtained by subtracting the camera motion-induced flow field from the combined flow field.
Pseudo-labels are generated through adaptive thresholding techniques based on the magnitude of estimated IMO motion field. Running inference \ours{} is simple without parameter turning because while the training process requires geometry-based labels, only events are used for prediction. 
Unlike many previous works, we do not assume simplified motion models or a  known number of objects.

\section{Related Work}
\begin{figure}[tb]
    \centering
    \includegraphics[width=\linewidth]{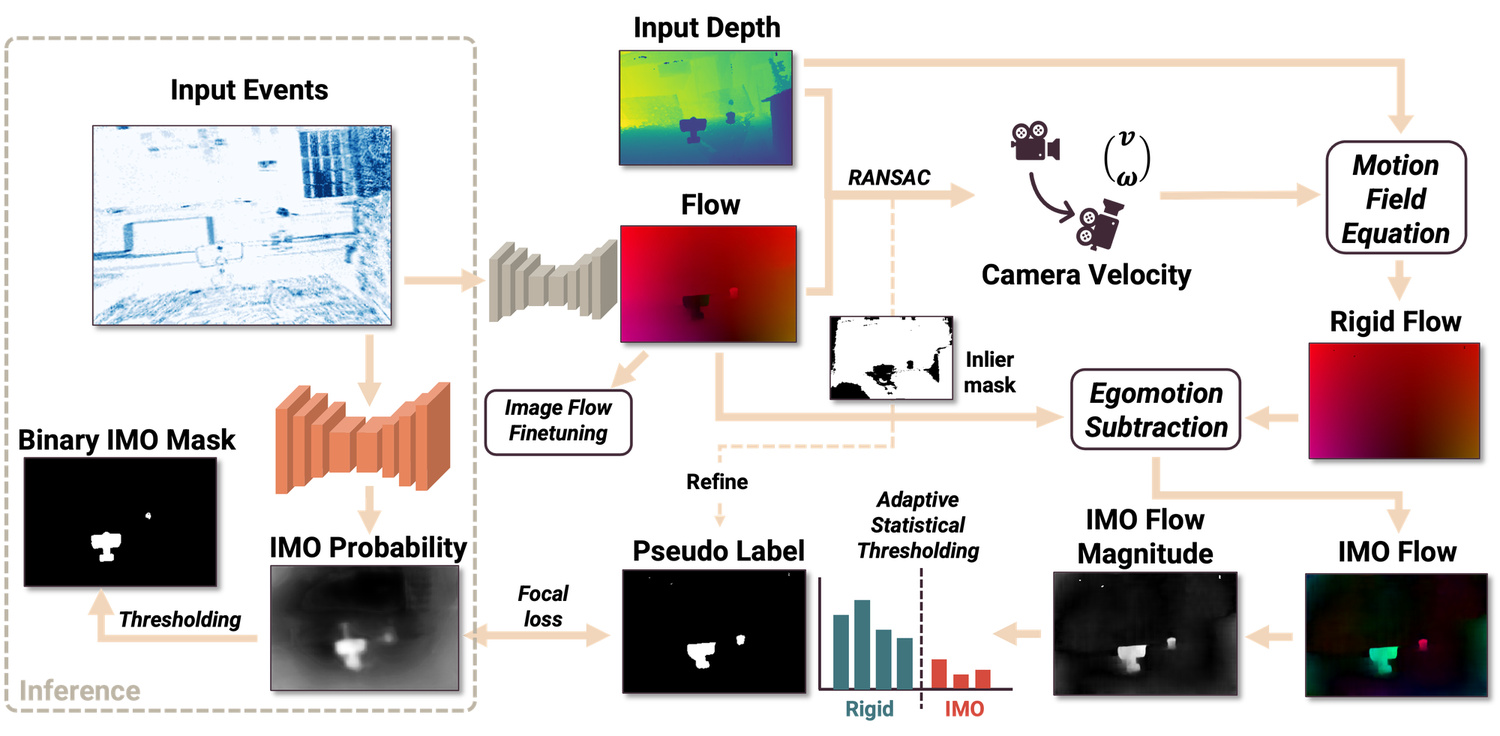}
    \caption{Proposed pipeline. \textbf{Left Dotted Box}: we train a network to directly predict IMO masks from events. \textbf{Rest of Figure}: we use a geometric self-labeling method to generate IMO pseudo-labels for supervision. Our framework uses off-the-shelf optical flow (fine-tuned on image-based flow) and input depth. The camera motion fitted from flow and depth through RANSAC is used to compute egomotion flow. Pseudo-labels are generated through adaptive thresholding techniques based on the magnitude of estimated IMO motion field. 
    We take the best of both worlds of deep learning and optimization: \textbf{1)} simple and robust inference with a simple feed-forward pass, and \textbf{2)} scalable with no expensive annotations required to train the network.}
    \vspace{-0.5cm}
    \label{fig:pipeline}
\end{figure}
\subsection{Event-based Motion Segmentation}
Recent advances in event-based motion segmentation research are driven by several event-based datasets. EVIMO~\cite{mitrokhin2019ev} is a motion segmentation data set that contains more than 30 minutes of various motions of scanned objects with a moving camera. Objects are geometrically tracked with a multi-camera tracking system (Vicon) and then projected onto a tracked camera. In the EVIMO paper, a baseline approach has been proposed to learn the mixture of unsupervised 3D velocities, depth, and flow from events. Motion segmentation is trained using the motion masks provided in the datasets on top of the learned mixture weights. Recently, Burner et al. released EVIMO2~\cite{burner2022evimo2}, which uses VGA resolution cameras. Evdodgenet~\cite{sanket2020evdodgenet} predict camera velocity by deblurring ground events using a downward-facing event camera and a motion segmentation network to identify objects that need to be dodged.
Stoffregen et al.~\cite{stoffregen2019event} proposed an Expectation-Maximization framework that assigns events to different motion clusters by optimizing the event-based contrast maximization. EMSGC~\cite{zhou2021event} is an optimization method that uses a graph cut method to cluster events in the x-y-t event space based on parametric flow. Mitrokhin et al.~\cite{mitrokhin2020learning} use a graph neural network to learn the segmentation masks directly in the event point space. GConv~\cite{mitrokhin2020learning} uses a graph neural network to learn event-based segmentation on graphs constructed on down-sampled events. SpikeMs~\cite{parameshwara20210} apply a spiking neural network (SNN) architecture  that allows incremental updates of the prediction over a longer time horizon. We compare the features of these methods with our work in Table~\ref{tab:feature}.
\subsection{Unsupervised Motion Segmentation}
Motion estimation and segmentation are coupled problems~\cite{ranjan2019competitive}.
In classical computer vision,  motion segmentation is  solved by  optimization  that simultaneously estimates parametric flow and  motion labels. Early layered flow models~\cite{darrell1991robust, ju1996skin, kumar2005learning} model the flow field as multiple motion layers, each representing a parametric motion field. To robustly optimize the different flow patterns, mixture flow models are proposed to compose the overall optical flow field with multiple simpler parametric flow fields. These methods usually assume a fixed number of clusters and simplified parametric forms of the individual flow component. Later, several works have found that clustering the orientation of the flow field leads to good segmentation results~\cite{bideau2016s, narayana2013coherent}. 

These problems have been significantly improved with the advancement of neural networks, which provide the ability to learn motion and structure prior from a large amount of data. 
The most common way to approach the problem of estimating ego-motion is to directly predict flow, depth, and egomotion~\cite{chen2019self, ranjan2019competitive, yin2018geonet, zou2018df}. These quantities are related by the rigid motion field equation, and thus, geometric constraints can be used for joint optimization to improve overall performance. Zhu et al.~\cite{zhu2018robustness} inserted a nondifferentiable RANSAC layer to allow explicit handling of nonrigid and/or independently moving objects in the scene. Casser et al.~\cite{casser2019depth} modeled both camera ego-motion and objects motion model in 3D space; however, the 3D object motion estimator requires precomputed semantic segmentation masks as input, which are unavailable in most settings. 

The incompatibility between independent motion and camera motion also creates opportunities for segmentation. Ranjan et al.~\cite{ranjan2019competitive} proposed an adversarial collaboration framework to explain and assign pixels to IMO or rigid backgrounds. Furthermore, informatic-theoretic approaches have been proposed to supervise segmentation networks by training an inpainter and a segmenter~\cite{yang2019unsupervised}. 
The motion segmenter predicts a foreground mask so that the inpainter cannot recover the masked foreground region from the background. On the other hand, the inpainter tries to inpaint the flow field using a background flow pattern. These works tend to work better on datasets with relatively simple camera motion and a single IMO. 
Another line of approaches related to our work is geometric self-labeling. Yang and Ramanan~\cite{Yang_2021_CVPR} trained a network to segment objects based on the error in the flow of the predicted scene. Zheng and Yang~\cite{zheng2021rectifying} refined pseudo labels by examining the uncertainty of semantic segmentation. Xie et al.~\cite{xie2024moving} uses the Segment Anything Model (SAM) to assist with flow-based motion grouping.
\vspace{-0.2cm}

\section{Preliminaries}

In this section, we geometrically define Independently Moving Objects (IMOs) in a 2D motion field.  We consider the first-order instantaneous optical flow derived by Longuet-Higgins et al.~\cite{longuet1980interpretation}. For a point $P = (X, Y, Z)$ that is observed by a camera $C$ that moves instantaneously with linear velocity $v$ and angular velocity $\omega$, its 3D motion field is written as:
\begin{align}
   \Dot{P} &= -\mathbf{v} - \mathbf{\omega} \times \mathbf{P} =
-
\begin{bmatrix}
    v_x \\
    v_y \\
    v_z
\end{bmatrix}
-
\begin{bmatrix}
     0 & -\omega_z & \omega_y \\
     \omega_z & 0 & -\omega_x\\
     -\omega_y & \omega_x & 0
\end{bmatrix}
\begin{bmatrix}
     X \\
     Y \\
     Z
\end{bmatrix}.
\label{eqn:moving_point}
\end{align}
Assuming a pinhole camera model, the point $(X, Y, Z)$ is projected to $(\frac{X}{Z}, \frac{Y}{Z})$, whose derivative with respect to time is: \begin{align}
\begin{bmatrix}
     \Dot{x} \\
     \Dot{y} 
\end{bmatrix}
= \frac{1}{Z}
\begin{bmatrix}
     \Dot{X} \\
     \Dot{Y} \\
\end{bmatrix}
-
\frac{\Dot{Z}}{Z^2}
\begin{bmatrix}
     X \\
     Y
\end{bmatrix}.
\label{eqn:rigid_flow}
\end{align}
Plugging Equation~\ref{eqn:moving_point} into Equation~\ref{eqn:rigid_flow}, we obtain the 2D motion field generated from point $P$:
\begin{align}
\begin{bmatrix}
     \Dot{x} \\
     \Dot{y} 
\end{bmatrix}
=\frac{1}{Z}
&\begin{bmatrix}
    -1 & 0 & x \\
    0 & -1 & y 
\end{bmatrix}
\begin{bmatrix}
   v_X \\
   v_Y \\
   v_Z
\end{bmatrix}
+ \begin{bmatrix}
    xy & -(1 + x^2) & y \\
    1 + y^2 & -xy & -x
\end{bmatrix}
\begin{bmatrix}
   \omega_X \\
   \omega_Y \\
   \omega_Z
\end{bmatrix}.
\end{align}
It can be seen that for an object moving in the camera frame with linear and angular velocity $v_o$ and $\omega_o$, the combined motion field can be  written as the sum of two motion fields $\Psi(v_c, \omega_c, X, Y, Z)$ and $\Psi(-v_o, -\omega_o, X, Y, Z)$, as object velocity can be thought as the opposite of camera velocity. In the following sections, we slightly abuse the notation to write $\Psi(x)$ to indicate the motion field of a 2D point $x$ which inversely projects to point $[X, Y, Z]$ in the camera frame. More generally, with multiple IMOs, the motion field can be written as:
\begin{align}
\Psi(x) = \Psi_{cam}(x) + \sum_i \Psi_{O_i}(x) \mathbbm{1} [x \in O_i ],
\label{eqn:mix}
\end{align}
where $O_i$ represents the $i$th object in the scene, where $\cup_{i=1}^{n}O_{i}$ represents all independently moving points in the scene that can be observed in the camera. $n$ is the total number of objects. Since the objects are assumed to be non-transparent, for each point observed by the camera, only one object contains this point:
\begin{align}
    \cap_{i=1}^n O_i = \emptyset.
\end{align}
From Equation~\ref{eqn:mix}, it can be seen that the objects and the camera have independent motion patterns. It it worth noting that previous literature usually models this as a mixture model~\cite{layer1995layered} where the indicator function $\mathbbm{1}[x\in O_i]$ is replaced with a weight $w_i$ and the camera motion field is weighted by $w_{cam}$ such that $w_{cam} + \sum_i{w_i} = 1$. The weight $w_i$ is a soft weight that indicates the likelihood that a point belongs to an object $O_i$ or the camera. Similarly, Stoffregen et al.~\cite{stoffregen2019event}, Mitrokhin et al.~\cite{mitrokhin2019ev}, Zhou et al.~\cite{zhou2021event} all employed this mixture formulation to enable segmentation among several candidate motion models. Either an Expectation-Maximization frame is used to optimize the weights directly, or a network is used to learn the mixture weights.

However, several underlying assumptions are made here to reduce the generalization ability of such approaches. First, such mixture models assume a fixed number of candidate models to initialize. These values cannot be easily tuned and depend heavily on the scene. In our experiments, we find the number of clusters cannot easily be selected without knowing beforehand the number of objects in the test sequence. Second, the mixture model makes strong assumptions about the parametric motion model. EMSGC~\cite{zhou2021event} uses 4 to 12 parameter models on different scenes. EMMC~\cite{stoffregen2019event} uses linear, rotational, 4-DOF and 8-DOF models. The most general model is EVIMO~\cite{mitrokhin2019ev}, which uses translational-only models for the object and a full rigid motion field for the camera.

In comparison, we deploy the exact formulation in Equation~\ref{eqn:mix}, and estimate the IMO motion weights directly through a per-pixel classification network, utilizing a discriminative power of a neural network over a large amount of data. This choice leads to a major challenge in event-based research, which is the lack of labeled data. In the next sections, we explain how we train the network without labeled motion masks.

\section{Unsupervised Motion Segmentation}
\begin{figure}[tb]
    \centering
    \includegraphics[width=\linewidth]{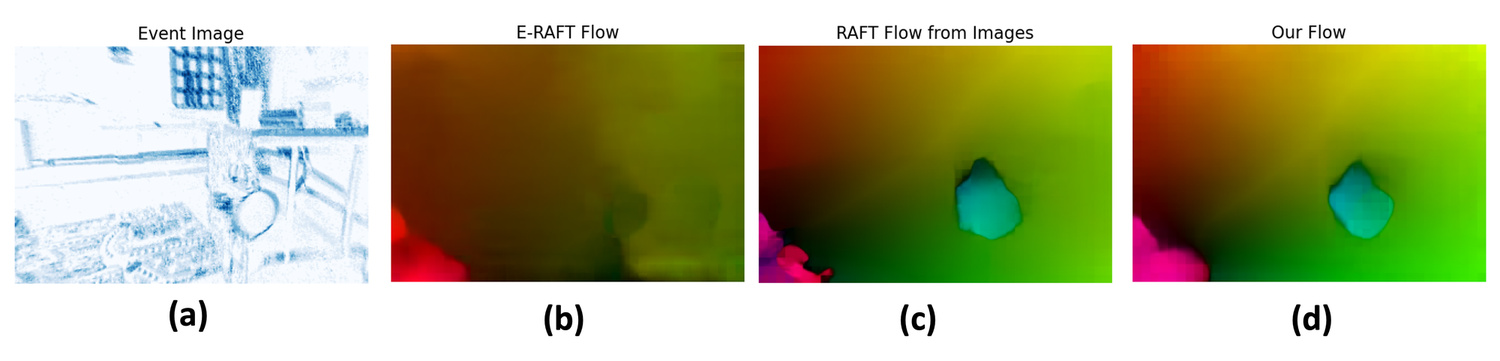}
    \caption{\textbf{(a)}: Events projected onto x-y space. \textbf{(b)}: E-RAFT flow. \textbf{(c)}: RAFT flow from Images. \textbf{(d)}: our optical flow containing independent motion. Independent motions are clearly missing from E-RAFT. Flow fields are predicted on the wall test sequence of EVIMO. The color indicated direction. Best viewed in color.}
    \label{fig:flow}
\end{figure}

\definecolor{tabfirst}{rgb}{1, 0.7, 0.7} %
\definecolor{tabsecond}{rgb}{1, 0.85, 0.7} %
\definecolor{tabthird}{rgb}{1, 1, 0.7} %

\begin{table}[tb]
    \centering
    \caption{Quantitative Evaluation on EVIMO. Event-masked IoU on predicted masks and gt masks is calculated as described in \textbf{Eqn.}~\ref{eqn:iou}. Our method compares favorably with EMSGC, which is the only one other than \ours{} that does not need labels. Our method performs competitively with other supervised methods. ``Baseline CNN" is our network-trained ground truth masks. EMSGC requires per-scene parameter tuning. For fair comparisons, we take the top 30 and 50 percent of EMSGC IoU.}
    \begin{tabular}{c  c  c   c  c  c}
    \toprule
        & \hspace*{0.4cm}Table\hspace*{0.4cm} & \hspace*{0.4cm}Box\hspace*{0.4cm} & \hspace*{0.4cm}Floor\hspace*{0.4cm} &\hspace*{0.2cm} Plain Wall\hspace*{0.2cm} & \hspace*{0.2cm}Fast Motion\hspace*{0.2cm} \\
    \midrule
    \multicolumn{6}{c}{\textbf{Supervised Methods}} \\
    \midrule
    Baseline CNN  & 66$\pm$23 & 50$\pm$23 & \textbf{74$\pm$13} & 60$\pm$20 & 52$\pm$24 \\
    Motion-blurred Video &24$\pm$25 & 28$\pm$30 & 40$\pm$25 & 30$\pm$26 & 14$\pm$18 \\
    EVIMO~\cite{mitrokhin2019ev}    & \textbf{79$\pm$6} & \textbf{70$\pm$5} & 59$\pm$9 & 78$\pm$5 & \textbf{67$\pm$3} \\
    EVDodgeNet~\cite{sanket2020evdodgenet} & 70$\pm$8 & 67$\pm$8 & 61$\pm$6 & 72$\pm$9 & 60$\pm$10  \\
    SpikeMS~\cite{parameshwara20210}  & 50$\pm$8 & 65$\pm$8 & 53$\pm$16 & 63$\pm$6 & 38$\pm$10 \\
    GConv~\cite{mitrokhin2020learning}    & 51$\pm$16 & 60$\pm$18 & 55$\pm$19 & \textbf{80$\pm$7} & 39$\pm$19 \\

    \midrule
    \multicolumn{6}{c}{\textbf{Unsupervised Methods}} \\

    EMSGC~\cite{zhou2021event}  Top 30$\%$  & \textbf{55$\pm$17} & 24$\pm$28  & 18$\pm$29 & 24$\pm$33 & 43$\pm$27  \\
    EMSGC~\cite{zhou2021event}  Top 50$\%$  & 36$\pm$27 & 14$\pm$25  & 11$\pm$24 & 15$\pm$28 & 26$\pm$29 \\
    \ours{} (Ours)   & 50$\pm$21 & \textbf{45$\pm$24} & \textbf{56$\pm$15} & \textbf{53$\pm$19} & \textbf{44$\pm$21}  \\
    \bottomrule
    \end{tabular}
    \label{tab:iou}
\end{table}
In Figure~\ref{fig:pipeline}, we show the pipeline of \ours{}. Generating motion labels on a large scale has a been a challenging problem. The most scalable solution is collecting data in simulation~\cite{mayer2016large, DFIB15}. In video datasets such as DAVIS16~\cite{Perazzi2016}, the motion masks of objects are usually labeled by humans. In driving datasets that have high accuracy depth sensros, such as KITTI~\cite{geiger2013vision}, IMOs are mostly cars. These objects are removed and inserted back using fitted car CAD models. In certain constrained cases, the labels can be generated by projecting known objects into the current camera frame. In EVIMO~\cite{mitrokhin2019ev}, the authors scanned the environment and objects before collecting dynamic motion. During data collection, VICON markers are attached to objects and cameras so that the relative poses between the camera, objects, and room are known. The object masks are then subsequently obtained by projecting the 3D model of the object onto the current camera. Despite this automatic labeling scheme, the amount of work required to calibrate the system and provide high-quality object scans makes this supervising method not transferable to general scenes.

In this section, we propose a framework for automatically obtaining labels taking advantage of the results of the CNN-based optical flow~\cite{zhu2018ev, zhu2019unsupervised, teed2020raft, gehrig2021raft, ye2020unsupervised, hamann2024motion, zhang2026match} estimation. The event-based optical flow networks are usually trained with large-scale event-based dataset~\cite{gehrig2021dsec, chaney2023m3ed, zhu2018multivehicle, zhu2021eventgan, shi2026evis}. Our method is based on geometric error rather than on the semantics of the objects, which allows it to be applied on a large scale. We explain how roughly accurate labels can be generated only using depth and camera data. In addition, we describe how we train a robust event-based motion segmentation network completely without human annotation. Our pipeline is mainly composed of two parts: a robust pseudo-label generation module and an event motion segmentation network. The data required for training is only the depth map in the camera frame. The depth information is only used during training in our geometry-based pseudo-label generation module. Such data are not required during inference. Instead, we train a per-pixel classifier that takes in events and produces a binary segmentation mask.

\subsection{Optical Flow with Independent Motion}
\begin{table}[tb]
    \centering
    \setlength{\tabcolsep}{10pt}
    \caption{Optical flow comparison. E-RAFT underperforms when there is independent motion. We report EPE metric as described in E-RAFT~\cite{gehrig2021raft}.}
    \begin{tabular}{c c c c c c}
    \toprule
     & Table & Box & Floor & Wall & Fast  \\
     \midrule
    E-RAFT~\cite{gehrig2021raft} & 11.150 & 14.902 & 4.983 & 8.036 & 20.471  \\
    Ours & \textbf{1.550} & \textbf{3.432} & \textbf{1.036} & \textbf{2.062} & \textbf{5.331}  \\
    \bottomrule
    \end{tabular}
    \label{tab:flow_comparison}
\end{table}
The high temporal resolution of the events preserves rich temporal information in x-y-t space, which allows robust estimation of optical flow under various challenging conditions. Early work achieves this estimation by plane fitting~\cite{benosman2013event}, which produces an event-based optical flow only on regions with events. EV-FlowNet~\cite{zhu2018ev} and E-RAFT~\cite{gehrig2021raft} are trained neural networks that learn the dense optical flow from events. In our formulation, it is critical to have dense flow predictions in order to compute the residual error between camera motion and the observed flow field. In this work, we used the E-RAFT flow network pretrained on DSEC. We fine-tuned the flow on the predicted flow from grayscale images using RAFT~\cite{teed2020raft}. In Figure~\ref{fig:flow}, we show examples of three types of optical flow. RAFT~\cite{teed2020raft} is the state-of-the-art optical flow method for images. E-RAFT extends the RAFT framework to events. It can be seen that our fine-tuned flow correctly estimates the flow for IMO objects. This is consistent with the discovery of Shiba et al. that E-RAFT performs poorly on independently moving objects~\cite{shiba2022secrets}. 
\paragraph{Optical Flow with Independent Motion}
Flow networks trained on driving data cannot be easily used for IMO detection. To show this, we compared our optical flow results with the state-of-the-art E-RAFT models pre-trained on DSEC~\cite{gehrig2021dsec}. For this evaluation, we used the architecture of E-RAFT as is and only fine-tune the flow based on image-based flow. Since the ground-truth optical flow of EVIMO is not provided, we supervised the high-quality optical flow computed using RAFT~\cite{teed2020raft} with photometric matching and refinement. In Table~\ref{tab:flow_comparison}, we compare our fine-tuned flow with pre-trained E-RAFT flow on unseen test sequences in EVIMO using RAFT flow as ground truth. In our experiments, we observe that the performance gap between our \ours{} flow network and E-RAFT is tightly correlated with the dynamic of the scene. In our experiments, due to the missing IMOs, the E-RAFT baseline cannot provide good pseudo-labels for training the downstream network.
\begin{figure}[tb]
    \centering
    \includegraphics[width=\linewidth]{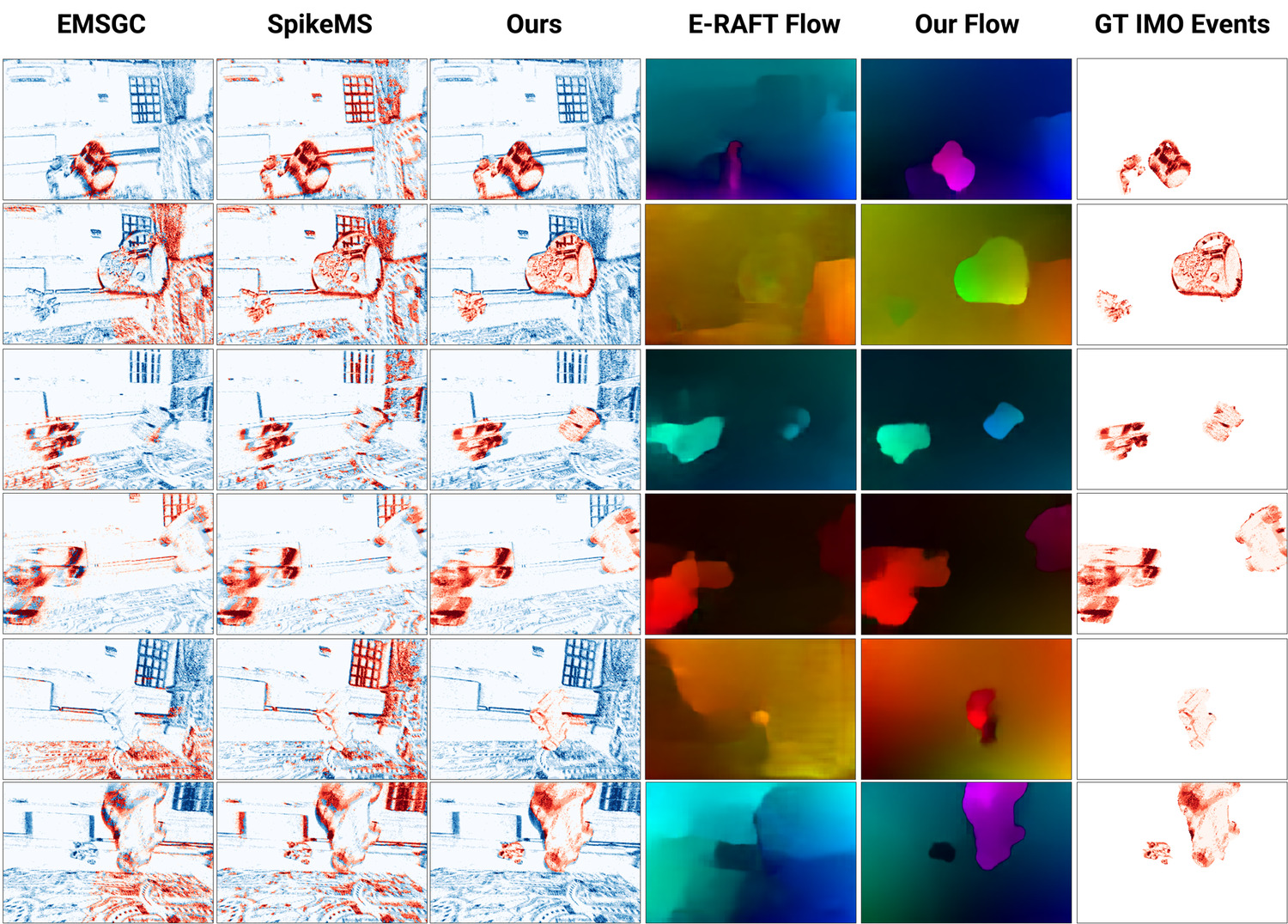}
    \caption{\textbf{Columns 1 to 3}: Segmentation results of EMSGC, SpikeMS, \ours{} and EVIMO-Supervised. \textbf{Columns 4 to 5}: E-RAFT flow output (trained in DSEC) and our fine-tuned flow network. \textbf{Column 6}: Segmented IMO event using ground truth. It can be seen that \ours{} produces sharper and more consistent masks than the baseline methods.  }
    \label{fig:quality}
\end{figure}
\subsection{Robust Camera Motion Estimation}
Traditionally, the motion segmentation problem can be seen as a chicken-and-egg problem because IMOs can significantly bias camera motion estimation if they are not properly filtered. Several self-supervised methods for joint motion estimation approaches are susceptible to this problem. For example, Zhu et al.~\cite{zhu2019unsupervised} jointly learned egomotion, depth, and flow assuming rigid scenes, which is dependent on a network to ignore independent motions. E-RAFT~\cite{gehrig2021raft}, although it does not learn ego-motion directly, has been shown to underperform in independent motion regions~\cite{shiba2022secrets}. Thus, a robust camera motion module needs to be designed to avoid further blurring of the decision boundary between IMO motion and camera motion. To this end, we take advantage of the classical outlier rejection techniques and use Random Sample Consensus (RANSAC) to estimate camera motion. In general, RANSAC is used to solve the following problem:
\begin{align}
\theta = \argmin_{\theta} 
\sum_{i=1}^N \rho(\epsilon(u_i; \theta)),
\end{align}
where $\epsilon$ is an error function, $\rho$ is a robust likelihood function, $N$ is the total number of observations, and $u_i$ is the observed motion field at pixel $i$ with respect to the camera motion given the velocity $\theta$. We notice that the error term $\epsilon(u_i; \theta)$ corresponds exactly to $\sum_i \Psi_{O_i}(x) \mathbbm{1} [x \in O_i ]$, the second term in Equation~\ref{eqn:mix}. A naive optimization without outlier rejection will bias the motion estimation towards the motion of near and fast-moving objects.
Based on Equation~\ref{eqn:rigid_flow}, the camera motion $(v_x, v_y, v_z, \omega_x, \omega_y, \omega_z)$ can be solved by the linear equation:
\begin{align}
\begin{bmatrix}
    -1/z_1 & 0& x_1/z_1& x_1y_1& -(1+x_1^2)& y_1 \\
         &&&\vdots \\
         \-1/z_n & 0& x_n/z_n& x_ny_n& -(1+x_n^2)& y_n \\
         0& -z_1^{-1} & y_1/z_1& 1+y_1^2& -x_1y_1& -x_1 \\
         &&&\vdots \\
         0& -z_n^{-1} & y_n/z_n& n+y_n^2& -x_ny_n& -x_n
\end{bmatrix}
\begin{bmatrix}
    v_x \\
    v_y \\
    v_z \\
    \omega_x \\
    \omega_y \\
    \omega_z \\
\end{bmatrix}
=
\begin{bmatrix}
    \dot{x_1} \\
    \vdots \\
    \dot{x_n} \\
    \dot{y_1} \\
    \vdots \\
    \dot{y_n}
\end{bmatrix}
,
\label{eqn:lin_eq}
\end{align} 
where $z_i$, $x_i$, $y_i$ are the depth values (input) and the pixel coordinates of the $i$th pixel and $(\dot{x_i}, \dot{y_i})$ is the calibrated optical flow from events. We sample 3 points every time to solve the equation for a maximum of 300 iterations, or a stop probability of 0.999 is reached. Then we use all inlier pixels to solve the over-constrained least square problem using SVD. We present the quantitative pose estimation results in Table~\ref{tab:pose_eval}. Our average translational error in relative pose estimation is sub-centimeter in Table and Floor sequences. The error is 4 for the extremely challenging Fast sequence. This shows the robustness of our pose tracking method that fuses an accurate flow method with robust geometric estimation method.
\begin{table}[tb]
    \centering
    \setlength{\tabcolsep}{10pt}
    \caption{Relative camera pose estimation using flow displacement. The translational error is defined as the mean squared error between the estimated and ground truth camera positions. The rotation error is defined as $log_m(R_{gt}^T R_{pred})$.}
    \begin{tabular}{c c c c c c}
    \toprule    
    &  Table & Box &  Floor  & Wall & Fast\\
    \midrule
    Trans. (m) &  0.0082 & 0.0251 &  0.0075 & 0.0141 & 0.0416 \\
    Rot. (rad) &  0.0348 & 0.0412 &  0.0261 & 0.0296 & 0.1110\\
    \bottomrule
    \end{tabular}
    \label{tab:pose_eval}
\end{table}

\subsection{Adaptive Geometry-based Thresholding}
We combine accurate flow estimation from events and robust motion estimation to produce a residual flow field. In contrast to model-based approaches in previous event-based motion segmentation works, we do not assume a fixed number of parametric flow models. In Section~\ref{sec:exp}, we show failure cases of parametric flow due to the high variation of motion and depth in real data. Since no competing models are learned or optimized, selecting an appropriate threshold for the magnitude of the residual flow becomes a crucial step. In analyzing the data, we find that the error usually demonstrates a bimodal distribution, where one peak corresponds to the correct rigid motion, and the other model concentrates at a much higher mean. Since there is usually no fixed threshold value due to the variation of noise and depth, we adopt a statistically robust thresholding method based on Otsu's method~\cite{otsu1979threshold}.

Given a set of pixels $\Lambda = \{q_i\}$, the residual flow function for each pixel is predicted by computing the $l^2$ norm of the residual flow: $r(q_i) = ||\Psi(q_i) - \Psi_{cam}(q_i)||_2$. Modeling the residual $r(q_i)$ as a bimodal distribution, choosing a threshold $\hat{r}$ is treated as the problem of maximizing the variance  between the two classes. The two classes, by definition, are rigid areas and IMO areas. IMO areas have higher residual flow because they have different velocities than the camera. The problem can be solved efficiently with a simple 1D search if we define $R = \{r_j\}$ as the set of candidate solutions. The objective of the search is
\begin{align}
&\argmax_{r_j \in R} \sum_{k=0}^{r_j} P_k (\mu_bg(r_j) - \mu)^2 + \sum_{k=r_j}^{K_{max}} P_k (\mu_{imo}(r_j) - \mu)^2 \\
&\mu = \sum_{k=0}^{K_{max}} P_k k \quad \mu_{bg}(r_j) = \sum_{k=0}^{r_j} P_k k \quad \mu_{imo}(r_j) = \sum_{k=r_j}^{K_{max}} P_k k.
\end{align}
$P_k$ is the probability that a pixel $q_i$ falls into the bin $k$. We use 256 bins for this problem, and the histogram is clipped at 10 pixels. In our search, we applied a two-stage filter on Otsu's thresholding results. First, we examine the total variance of the histogram of errors; If the variance is greater than some threshold $\epsilon_{var}$, we do not look at this slice of events, since the flow prediction does not provide clear boundaries of the objects. Similarly, we compute the variance  between IMO pixels and BG pixels, based on the selected threshold $r_j$ and remove the training example if this value is too small. These two calculated variance values can be seen as a measure of confidence in the labels. Selecting confident labels is a crucial step in pseudo-label selection.

\subsection{Optional Depth Input}
To compute the optical flow, we take an optional depth map as input in Equation~\ref{eqn:rigid_flow}. In practice, the depth map can be acquired with a paired sensor or monocular depth network. This depth map is \textbf{only used in training} for generating the pseudo labels and \textbf{never used during actual inference} after the network has been trained. Alternatively, we can use parametric flow independent of the depth.
In \cite{mann1997video}, several parametric depth models are proposed. We take the 12-DOF biquadratic flow as an example. The flow for each pixel is defined as:
\begin{align}
& x^{\prime}=q_{x^{\prime} x^2} x^2+q_{x^{\prime} x y} x y+q_{x^{\prime} y^2} y^2+q_{x^{\prime} x} x+q_{x^{\prime} y} y+q_{x^{\prime}} \\
& y^{\prime}=q_{y^{\prime} x^2} x^2+q_{y^{\prime} x y} x y+q_{y^{\prime} y^2} y^2+q_{y^{\prime} x} x+q_{y^{\prime} y} y+q_{y^{\prime}},
\end{align}
The estimation of the camera-induced camera motion can be modeled as follows:
\setcounter{MaxMatrixCols}{20}
\begin{align}
\begin{bmatrix}
    x_1^2 & x_1y_1 & y^2_1 & x_1 & y_1 & 1 & 0 & 0 & 0 & 0 & 0 & 0 \\
         &&&&&&\vdots \\
    x_n^2 & x_ny_n & y^2_n & x_n & y_n & 1 & 0 & 0 & 0 & 0 & 0 & 0 \\
    0 & 0 & 0 & 0 & 0 & 0 & x_1^2 & x_1y_1 & y^2_1& x_1 & y_1 & 1 \\
         &&&&&&\vdots \\
    0 & 0 & 0 & 0 & 0 & 0 & x_n^2 & x_ny_n & y^n_n& x_n & y_n & 1 \\
\end{bmatrix}
\begin{bmatrix}
q_{x' x^2} \\
q_{x' x y} \\
q_{x' y^2} \\
q_{x' x} \\
q_{x' y} \\
q_{x'} \\
q_{y' x^2} \\
q_{y' x y} \\
q_{y' y^2} \\
q_{y' x} \\
q_{y' y} \\
q_{y'}
\end{bmatrix}
=
\begin{bmatrix}
    \dot{x_1} \\
    \vdots \\
    \dot{x_n} \\
    \dot{y_1} \\
    \vdots \\
    \dot{y_n}
\end{bmatrix}.
\label{eqn:12dof_flow}
\end{align} 
The estimation problem can be solved with RANSAC with six points. We present these results in Table~\ref{tab:abl_results} as ablation studies to provide how much the choice of flow modeling affects the segmentation performance.

\subsection{Event-based Motion Segmentation Network}
It can be seen from our pseudo-label generation framework that the task of independent motion segmentation can be seen as a combination of global and local motion estimation. As previously studied in the event-based flow literature~\cite{zhu2018ev, gehrig2021raft, zhu2019motion}, it is preferred to preserve motion information in events. For this purpose, we use the event volume representation, which encodes the temporal domain as discretized channels of a 3D tensor. A bi-linear interpolation kernel($k_b$) is used to distribute events to discretized bins based on their spatio-temporal proximity with these bins. We use the volume of events, which has been shown to be effective in understanding motion, as described in~\cite{zhu2019unsupervised, wang2022ev, wang2022evac3d, wang2025humanmotion, wang2025decompression}:
$E(x, y, t) = \sum_i p_i k_b(x-x_i)k_b(y-y_i)k_b(t-t_i^*)$.
We use 15 channels for the event volume to allow the network to extract fine temporal information from events. We provide details on the implementation of the network and the loss functions of~\ours{}. Our trained prediction module is a UNet-like convolutional neural network. The bottleneck layers facilitate the aggregation of global features, since the segmentation problem relies not only on the local flow pattern of events but also on the global motion pattern caused by the camera. We use a pre-trained ResNet34~\cite{he2016deep} encoder with pre-trained weights on ImageNet~\cite{deng2009imagenet}. Since objects usually occupy much less space than the rigid background, we use a Focal Loss~\cite{lin2017focal} to handle the class imbalance problem. The network is trained with an Adam optimizer using a learning rate of $2e-4$ on EVIMO Table, Wall, Floor, Box, and Fast training sequences.

\begin{figure*}[tb]
    \centering
    \includegraphics[width=\textwidth]{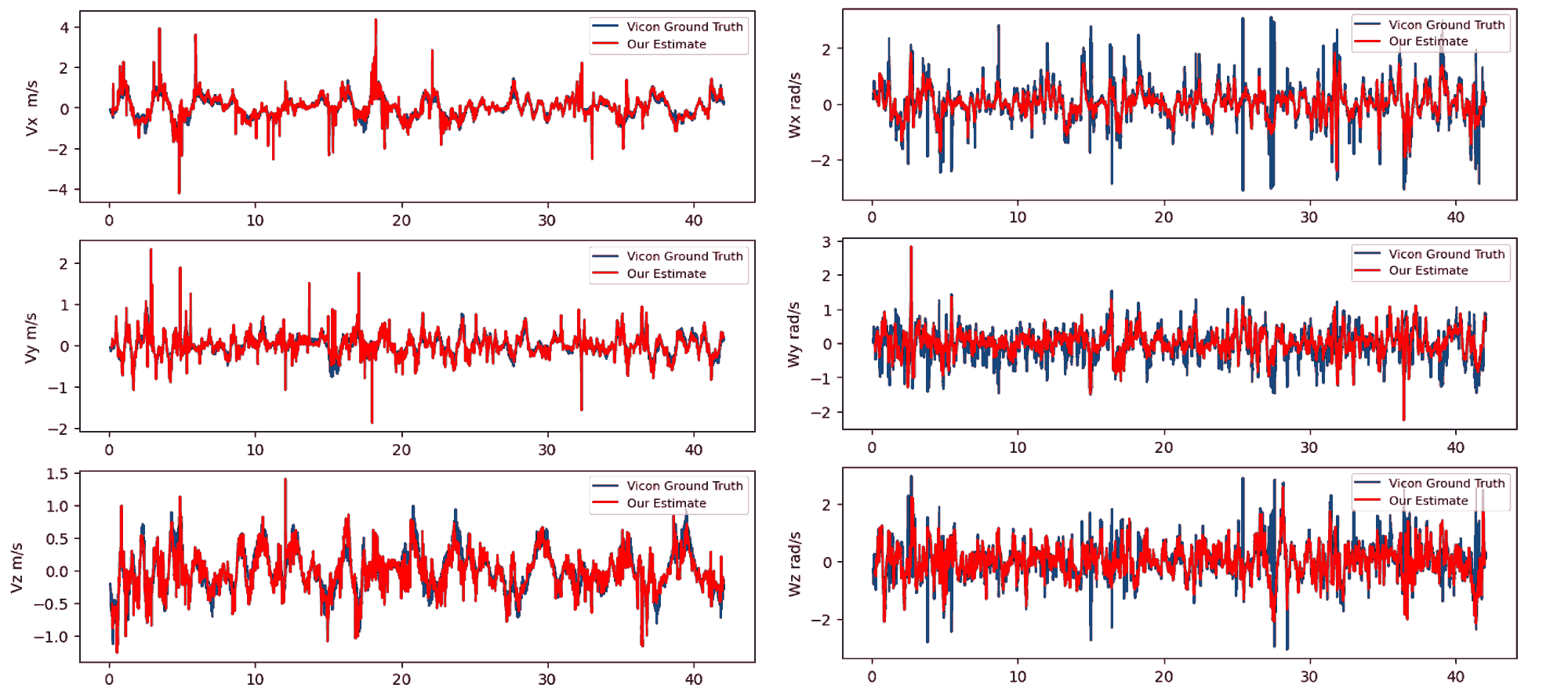}
    \caption{Estimated camera rotation from estimated optical flow. The results are shown for the whole evaluation sequence $wall\_00$. Best viewed in color. The left and right columns show translational and rotational error respectively.}
    \label{fig:camera_motion}
    \vspace{-.25cm}
\end{figure*}

\section{Experiments}
\label{sec:exp}

\textbf{Quantitative Evaluation}
In Table~\ref{tab:iou}, we report the IoU our \ours{} against competing methods on different classes of EVIMO. The IoU is computed on masked events directly in order to compare with single-event labeling approaches. The IoU score is computed as:
\begin{align}
    \text{IoU}(O_t, P_t, E_t) = \frac{ |(E_t \cap P_t) \cap (E_t \cap O_t)|}{ |(E_t \cap P_t) \cup (E_t \cap O_t)|},
    \label{eqn:iou}
\end{align}
where $E_t$ is the set of projected events surrounding time $t$. $P_t$ and $O_t$ are the projected mask and ground truth in 2D. $E_t, P_t$, and $O_t$ are all subsets of all pixels.
The comparison is evaluated at 40Hz, which is the default evaluation frequency for the dataset. Comparison methods can be divided into two classes: supervised and optimization-based. In supervised methods, a mask of a moving object is provided at each time. On the other hand, EMSGC in the table is an optimization-based method, which does not use mask labels. Instead, multiple motion models are fitted to the events by alternating between contrast maximization and flow fitting. It is worth noting that the EMSGC method is sensitive to parameters such as the class of the parametric model and the number of objects. 
Due to the large amount of evaluation data (thousands of frames per sequence), we were unable to tune the parameters for each slice. Instead, we tuned for each sequence and generously reported the top 30 performance to give it a fair comparison. This further emphasizes the advantages of our pseudo-label-based method over per-event-slice optimization. Please refer to the supplemental material for more details.
\begin{table}[tb]
    \centering
    \caption{Analysis of data processing (Pre), algorithm running (Run), and total time.}
    \begin{tabular}{c c c c c c}
    \toprule
        &  \textbf{Ours} &  \textbf{SpikeMS~\cite{parameshwara2021spikems}} & \textbf{GConv~\cite{mitrokhin2020learning}} & \textbf{EVIMO~\cite{mitrokhin2019ev}} & \textbf{EMSGC~\cite{zhou2021event}} \\
    \midrule
       Pre (ms) &  \textbf{3.35} & 10.56 & 698.62 & 16.74 & 33.38 \\
       Run (ms) &  \textbf{3.22} & 110.01 & 16.00 & 4.29 &  9496.04 \\
    \midrule
       Total (ms) & \textbf{6.57} & 120.57 & 715.62 & 26.85 &  9529.42\\
    \bottomrule
    \end{tabular}
    \label{tab:computation_time}
\vspace{-.4cm}
\end{table}
Our model outperforms the supervised spiking method and unsupervised ESMGC (with per-sequence tuning). It can be seen that our method is comparable to supervised methods on tables, floor, wall, and fast motion. Compared to supervised methods, the main disadvantage of our approach is the lack of sharp boundaries in prediction because the network is trained with noisy labels. To demonstrate the difficulty of the task when using frame cameras in low-light conditions, we include experiments with synthetic motion-blurred videos. We used SuperSlowMo to upsample videos to 640 fps and averaged frames to synthesize motion-blurred videos with a 0.125-second shutter time. We trained a supervised network on regular videos and tested it with motion-blurred videos.

\textbf{Qualitative Evaluation}
In Figure~\ref{fig:quality}, we provide qualitative examples of competing methods on the Wall sequence of the evaluation set. We show examples using methods whose source code is available. It can be seen that qualitatively, our results are very similar in quality compared with supervised CNN methods, largely outperform optimization-based methods, and even outperforms supervised SNNs. SpikeMS tends to sparsify the events and keep edges. EMSGC needs extensive tuning to get reasonable results. However, it still misclassifies IMO as rigid areas. With these noise predictions across the image from SpikeMS and EMSGC, IMO cannot be easily detected and handled, while our network produces spatially consistent segmentations.

\textbf{Computational Speed}
\ours{} trains a single feed-forward U-Net for IMO segmentation. Therefore, no heavy optimization is needed. In Table~\ref{tab:computation_time}, we show the computational time comparison between our method and the baseline methods. GConv~\cite{mitrokhin2020learning} uses a fast network, but the graph building operation takes a significant amount of time. EMSGC~\cite{zhou2021event} is faster in building the graph, but the per-event-slice optimization is extremely slow and is not guaranteed to converge. Our method shifts the computational burden to computing the pseudo-labels and enable single forward pass during inference. Our compute platform is a single RTX 3080 mobile GPU with an 8-core cpu.
\vspace{-.3cm}

\section{Ablation Studies}
We provide two ablation results, as shown in Table~\ref{tab:abl_results}. First, we compare the IMO labels computed with the un-refined flow directly predicted from E-RAFT. The off-the-shelf flow network is trained on DSEC, which has limited independent motion. Second, we show two results using parametric flow models (\textbf{b} and \textbf{c}) based on the definitions defined in previous work~\cite{mann1997video}. The parametric flow models provide the possibility without the input depth. With depth models with fewer degrees of freedom, the IMO labeling scheme still outperforms EMSGC. However, the best performance so far is the full model with input depth, which maximizes the IMO pseudo labeling quality. We leave the elimination of these dependencies to our future work.
\begin{table}[tb]
    \caption{\textbf{Ablation Studies}. (a) refers to using unrefined pre-trained E-RAFT flow network. (b,c) shows results using parametric flow models described in~\cite{mann1997video}.}
    \centering
    \setlength{\tabcolsep}{10pt}
    \begin{tabular}{c c c c c c}
    \toprule
              & Table & Box & Floor & Wall & Fast \\
    \midrule
        (a)ERAFT &  32$\pm$23 & 28$\pm$21 & 35$\pm$19& 42$\pm$22 & 27$\pm$23\\
        (b)6-DOF &  43$\pm$26 & 42$\pm$25 &  51$\pm$21 & 47$\pm$23 & 37$\pm$24\\
        (c)12-DOF &  47$\pm$24 & 40$\pm$25 &  56$\pm$18 & 49$\pm$22 & 37$\pm$25\\
    \midrule
        Ours  & \textbf{50$\pm$21}   & \textbf{45 $\pm$ 24}&  \textbf{56$\pm$15} & \textbf{53$\pm$19} & \textbf{44$\pm$21} \\
    \bottomrule
    \end{tabular}
    \label{tab:abl_results}
\end{table}

\section{Failure Cases and Limitations}
In Figure~\ref{fig:failure}, we show one false positive and one false negative output from our approach. Due to the extreme dynamic nature of the dataset, the residual between the background flow and the IMO flow is small. In particular, there are cases where the objects have near-zero velocities. These objects should be segmented if we consider its past motion, but should be excluded if we only look at current motion. This leads to the lack of temporal consistency in the prediction. A possible solution is adding constraints between the current IMO mask and the immediate past IMO masks during training. This could be applied during the pseudo-label generation phase too for better ground truth. In Figure~\ref{fig:failure2}, we demonstrate that adding temporal consistency can be helpful. The network lost track of the IMO at time, but it should know that an IMO is nearby by looking at the previous several mask predictions. A discontinuity in prediction should be penalized because the motion of an object can be seen as continuous in the events.

\begin{figure*}[tb]
    \centering
    \begin{subfigure}{0.4\linewidth}
        \centering
        \includegraphics[width=\linewidth]{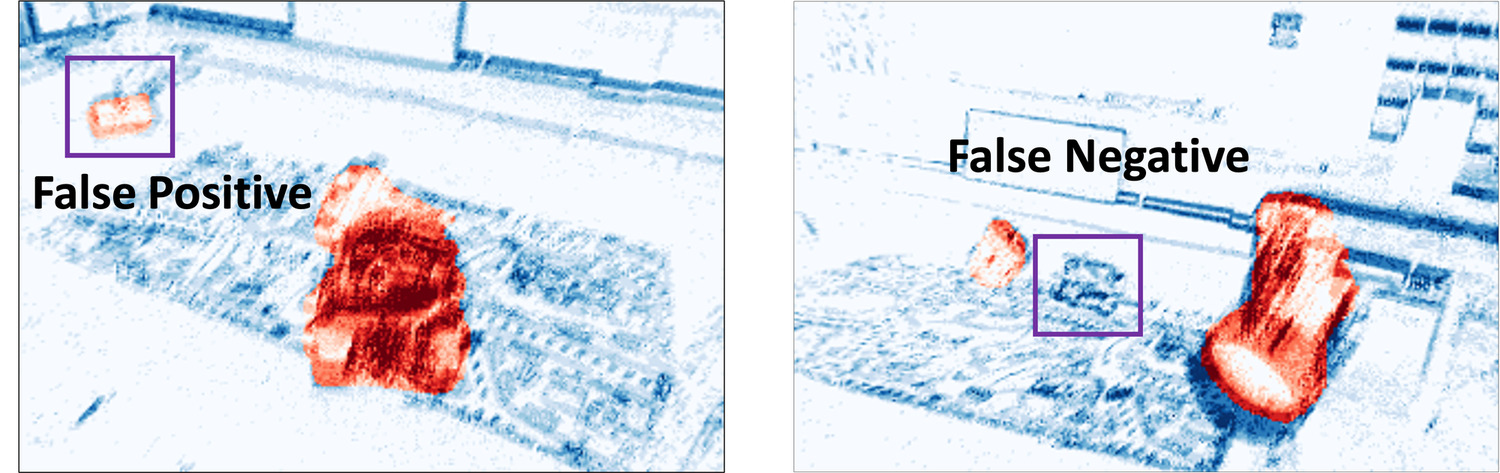}
        \caption{Failure cases of our method. On the left, the network incorrectly classifies a static square pattern on the ground as IMO. On the right, the network fails to find the apparent IMO in the scene. }
        \label{fig:failure}
    \end{subfigure}
~
    \begin{subfigure}{0.58\linewidth}
        \centering
    \includegraphics[width=\linewidth]{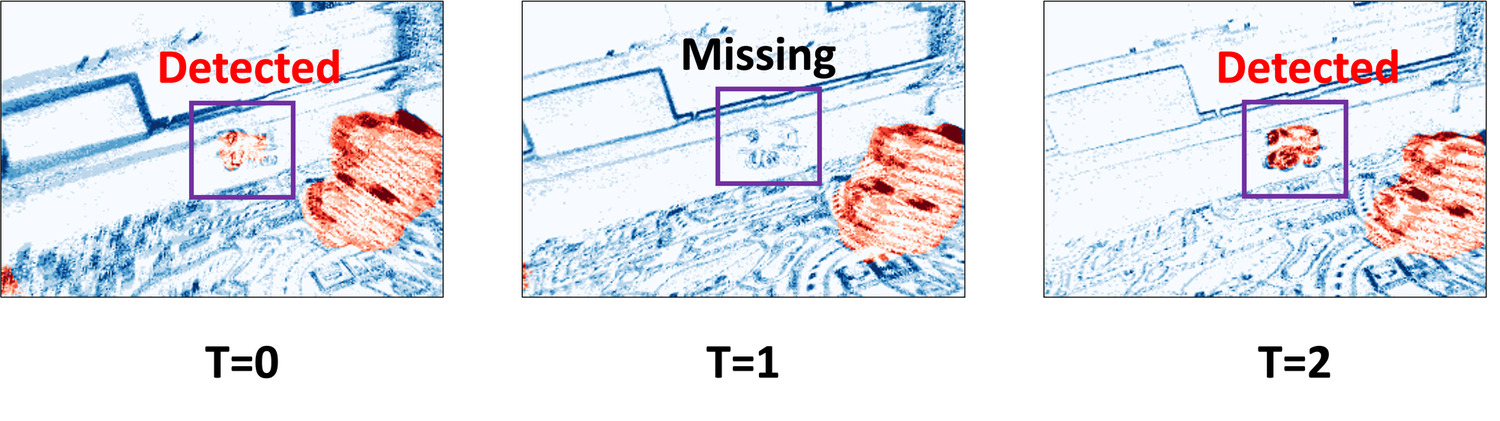}
    \caption{IMO predictions at three consecutive event slices. Our IMO detection runs on single slices of events. Occasional erroneous predictions do not have temporal consistency with the previous and next predictions.}
     \label{fig:failure2}
    \end{subfigure}
\end{figure*}

\section{Conclusion}
In this work, we tackle the problem of event-based segmentation from a geometric point of view. We focus on the major problem of event-based motion segmentation, which is the lack of labeled segmentation masks. Instead of using clustering techniques that require a fixed number of clusters and simplified parametric flow, our approach is purely geometric and robust to unseen semantic classes. Using the accurate event-based optical flow, we generated pseudo-labels based on the residual flow field defined by the difference between the estimated ego-motion field and the general motion field.  Ego-motion field was predicted using depth and a pre-trained flow network. With experiments on the EVIMO dataset, we show that our framework can be used to train downstream motion segmentation to perform competitively with supervised methods. 

\textbf{Acknowledgment}: We gratefully acknowledge the support by the following grants: NSF FRR 2220868, NSF IISRI 2212433, NSF TRIPODS 1934960, and ONR N0001422-1-2677

\bibliographystyle{splncs04}
\bibliography{main}

\fi %
\ifshowsupplementary
\setcounter{section}{0}

\clearpage
\setcounter{page}{1}

\title{\texorpdfstring{Supplementary Material: \\ 
Un-EVIMO: Unsupervised Event-based Independent
Motion Segmentation}
{}}
\author{}
\institute{}
\maketitle

\section{Additional Results}
\paragraph{Consecutive Segmentation Results}
In Figure~\textbf{3} of the main manuscript, we show samples of the test sequences. These images only show how individual predictions perform. In this supplementary material, we include more consecutive predictions to show that the network prediction is consistent, although the prediction at each time is independent. In Figure~\ref{fig:clips}, we show clips of continuous IMO segmentation to demonstrate temporal consistency. Similarly to our evaluation procedure, each image uses 0.025s of events. In each clip, we show six consecutive event slices in ascending order by time from left to right. We see in these figures that the boundaries of objects are sometimes misclassified as background events. There are two main reasons for this issue. First, the pseudo-masks are computed on a specific time rather than over a duration, which causes the network to predict the mask at a given time. Thus, the motion of objects during the time of the event slice can cause the network to underestimate the size of the IMO regions. In our experiments, networks trained with ground-truth labels also experience the same problem. Second, the sharp boundaries in the ground truth masks help the network learn better decision boundaries on binary classification. The baseline CNN we trained was able to keep slowly improving performance even after many epochs, whereas our method stopped improving after the first few epochs.

\paragraph{Egomotion Estimation Results}
 In the main manuscript, we assume that the camera pose can be accurately estimated from the flow prediction. Although we do not train a network to estimate the pose, accurate optical flow and depth can be combined to estimate egomotion robustly. In Figure~\ref{fig:pose}, we show the complete velocity estimate (linear and rotational) computed on unseen wall and floor sequences. Due to the high-frequency movements in EVIMO, our flow at 40Hz acts as a filter that smooths the velocities. On the other hand, the VICON ground truth is captured at 200Hz, which allows one to see the high-frequency vibrations. Overall, the robust RANSAC algorithm is able to estimate egomotion accuarately for idenfying the IMO regions.

\begin{figure*}
    \centering
    \includegraphics[width=\textwidth]{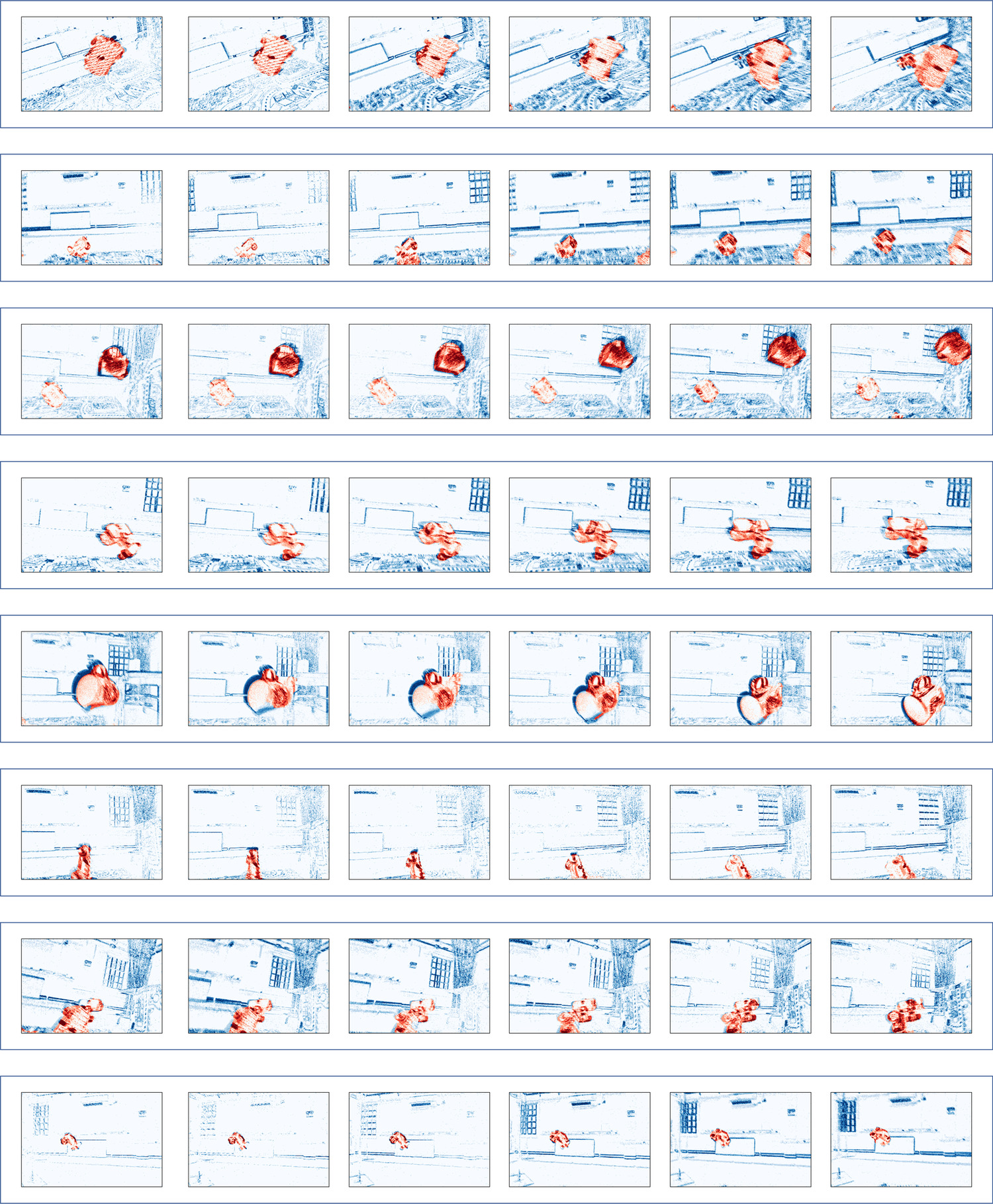}
    \caption{In each row, we show motion segmentation of a clip. Each clip shows temporally consistent segmentation results while each slice is predicted independently. Each row progresses temporally from left to right. Blue are background events and red are segmented IMO events. Best viewed in color.}
    \label{fig:clips}
\end{figure*}

\section{Additional Detection Rate Results}
In Section 5.1 in the main manuscript, we explain the lower performance of our approach compared to supervised baseline methods. Sharp boundaries in ground-truth masks provide stronger discriminative signals to the network. However, in fast motion estimation, an essential task is to locate the IMOs. In Table~\ref{tab:detect}, we computed the detection rate using the IoU threshold at $0.3$. 
\begin{table}[tb]
    \caption{Detection rate using IoU of 0.3 on all evaluation sequences on EVIMO~\cite{mitrokhin2019ev}.}
    \centering
    \setlength{\tabcolsep}{10pt}
    \begin{tabular}{c c c c c c}
        \toprule
         &Wall & Table & Floor & Box & Fast \\
         \midrule
         \shortstack{Detection Rate} &0.853 & 0.817 & 0.912 & 0.703 & 0.694\\
         \bottomrule
    \end{tabular}
    \label{tab:detect}
\end{table}
\begin{table}[tb]
    \caption{Full EMSGC evaluation results on all sequences. Each column corresponds to the top $K$ performance of EMSGC.}
    \centering
    \begin{tabular}{c | c c c c c c c c}
        \toprule
        \diagbox[]{Sequence}{Percentile} & K=0.3 & K=0.4 & K=0.5 & K=0.6 & K=0.7 & K=0.8 & K=0.9 & K=1.0  \\
        \midrule
        Table & 55$\pm$17 & 45$\pm$23 & 36$\pm$27 & 30$\pm$28 & 26$\pm$28 & 23$\pm$27 & 20$\pm$27 & 18$\pm$26 \\
        Wall & 24$\pm$33 & 18$\pm$31 & 15$\pm$28 & 12$\pm$26 & 11$\pm$25 & 9$\pm$23 & 8$\pm$22 & 7$\pm$21 \\
        Floor & 18$\pm$29 & 14$\pm$26 & 11$\pm$24 & 9$\pm$22 & 8$\pm$21 & 7$\pm$20 & 6$\pm$19 & 5$\pm$18 \\
        Fast & 43$\pm$27 & 33$\pm$29 & 26$\pm$29 & 22$\pm$28 & 19$\pm$27 & 16$\pm$26 & 15$\pm$25 & 13$\pm$24 \\
        Box & 24$\pm$28 & 18$\pm$26 & 14$\pm$25 & 12$\pm$23 & 10$\pm$22 & 9$\pm$21 & 8$\pm$20 & 7$\pm$19 \\
        \bottomrule
    \end{tabular}
    \label{tab:emsgc}
\end{table}

\section{Implementation Details}
\subsection{Data Preparation}
\begin{figure*}
\centering
\begin{subfigure}{.5\textwidth}
  \centering
  \includegraphics[width=\linewidth]{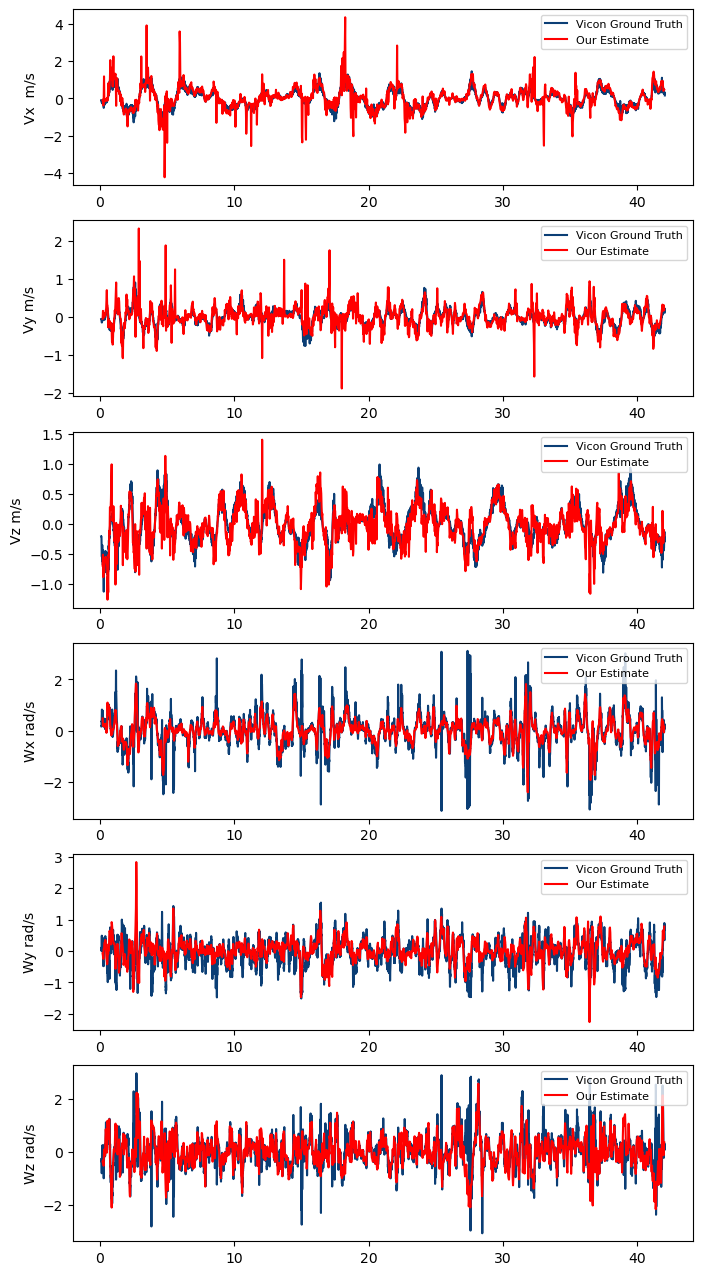}
  \caption{Results on Test Sequence Wall 00}
  \label{fig:sub1}
\end{subfigure}%
\begin{subfigure}{.5\textwidth}
  \centering
  \includegraphics[width=\linewidth]{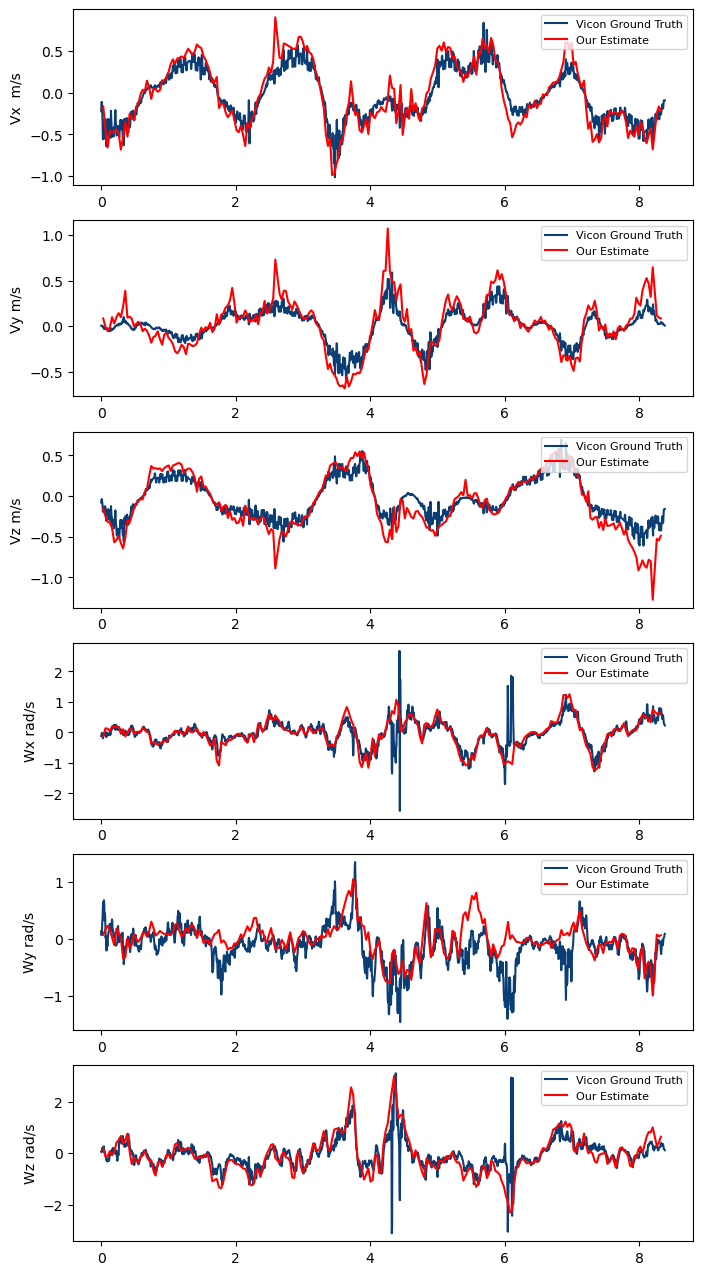}
  \caption{Results on Test Sequence Floor 01}
  \label{fig:sub2}
\end{subfigure}
\caption{Estimated linear and angular velocity in EVIMO evaluation sequences. Red is our estimated velocities from flow and RANSAC, and blue is the ground-truth velocity captured by VICON. It can be seen that the VICON estimates are at 200Hz, which is able to capture high-frequency motion more effectively, whereas our estimate is based on flow at 40Hz.}
\label{fig:pose}
\end{figure*}

As described in the main manuscript, we perform motion segmentation on events projected on $x, y$ space, allowing us to use existing image-based segmentation architectures. However, this does not imply that we discard time information from the input of the network. Instead, we use an event volume~\cite{zhu2019unsupervised} to encode the spatiotemporal information in the events. The input volume has a dimension of $(N, H, W)$, where $H$ and $W$ are the spatial dimensions of the event camera, and $N$ is the number of temporal bins used to discretize time. We use a relatively large number $15$ for $N$ to balance between the amount of temporal information and the usage of gpu. In EV-IMO~\cite{mitrokhin2019ev}, an DAVIS 346 is used for data collection, the sensor resolution is $260 \times 346$. In this dataset, a rather wide lens was used, which caused distortion. Our method assumes calibrated cameras, and thus we undistort the events and input depth and crop the images to $215 \times 320$. We use the raw resolution for training and inference. In addition, we clarify the training and test split of our network. Table, Wall, Floor and Box training sequences are used during training. We performed the test on all evaluation sequences from the same four classes. We perform an evaluation on all slices where at least a single object is present, when IoU is meaningful.

We notice that multiple modalities of the provided ground truth have built-in noise. For example, the depth maps are provided with holes and the scans have discontinuities on flat surfaces. Therefore, we only use depth maps up to 3 meters of the camera during training. In our pseudo-label generation, the holes in the depth map created discontinuous masks, which we use mathematical morphology to fill these holes. However, we find the network relatively robust to these changes because the pseudo-masks are themselves noisy. We report these engineering choices to ensure that the experiments are completely reproducible. 

\subsection{EMSGC Comparison}
EMSGC~\cite{zhou2021event} is an optimization-based method. We choose to compare with this method because it similarly does not use labeled training data. In this method, the authors propose to build a spatiotemporal graph and cut the graph based on contrast loss with respect to a predetermined number of motion models (2-parameter, 4-parameter, etc.). Like many optimization methods, EMSGC suffers from high sensitivity to hyperparameters. The exact hyperparameters for each sequence are not released with the code. These parameters include various motion models for the background and foreground, the weight $\lambda$ that balances local consistency versus spatial coherence, and MDL weight that determines how much we want to regulate the number of clusters. The details are in Section VI-C of the EMSGC paper~\cite{zhou2021event}, which states that the parameters are obtained based on properties of the data set and empirical tuning. However, in practice, it is difficult to know these parameters in advance, which weakens the method's ability to perform real-time inference.

In our initial tests, we used their open-source code and configuration files to run prediction on all evaluation sequences. However, this approach does not produce meaningful results in most of the event slices. Then, we tried tuning the parameters on each sequence separately, but found that per-sequence tuning was not sufficient for good performance. Due to the large amount of evaluation data (thousands of frames per sequence), we were unable to tune the parameters for each slice. Instead, we tuned for each sequence and used the highest $K$ percent of all IoU to compute the mean performance and then reported the results. The performance with low K value can be seen as an approximation of the upper-bound performance of the method. In Table~\ref{tab:emsgc}, we report the full results for selecting different $K$.

\subsection{SpikeMS Comparison}
For SpikeMS~\cite{parameshwara2021spikems}, we take quantitative results directly from their paper. However, there is a hyperparameter that specifies the maximum background-to-foreground ratio during evaluation. Therefore, the numbers reported in their paper can be seen as the upper bound of their performance. We used the pre-trained model released by the authors to generate the qualitative results. We notice that the network prefers to remove events in both IMO and background areas, which induces high recall, which works well in low background-to-foreground ratio scenarios. In our experiments with SpikeMS, the performance is significantly worse for general cases when the objects are smaller.

\subsection{Supervised CNN Baseline Comparison}
The original EVIMO network~\cite{mitrokhin2019ev} has a few auxilliary losses to assist segmentation. GConv~\cite{mitrokhin2020learning} uses a graph neural network on subsampled events where per-event labels are available. Comparing these methods does not give us a direct understanding of the effectiveness of the self-labeling mechanism. Therefore, we train a baseline network using the same architecture and ground truth labels. We report the results in Table 3 of the main article, labeled ``Baseline CNN". The average performance gap between this method and \ours{} is smaller than that between other listed methods. This simple baseline supports our hypothesis that our pseudo-labels are good approximation of the ground-truth labels, given that other factors have been controlled. We train the network using the same setting as the EVIMO network~\cite{mitrokhin2019ev}.

\subsection{Optical Flow Fine-tuning}
In EVIMO, only the flow of the foreground is given. We instead used RAFT~\cite{teed2020raft} to compute the optical flow from low-quality DAVIS images and use these as a good reference flow. We then fine-tuned the E-RAFT~\cite{gehrig2021raft} network for 10 epochs to allow E-RAFT to learn the IMO flow. In our experiments, we find that our flow network is able to overcome the missing IMO problem from this fine-tuning. In certain cases, it actually produces sharper flow than the RAFT flow labels. Since the ground-truth flow was missing from the general scene, we leave the full flow evaluation to future work. The fine-tuned network is forozen and is directly used as a fixed predictor in our pseudo-label generation module. We would like to emphasize that we do not claim new flow methods. Instead, we corrected the flow based on our need for accurate IMO motion estimation.

\subsection{Network Details}
In our experience with event data, pre-trained backbone usually gives the network better gradients for quicker convergence. We use a ResNet18 pre-trained on ImageNet as our encoder backbone. The event volumes are reshaped as $(15, 256, 256)$ via nearest neighbor interpolation and then fed into the network. The decoder is trained from scratch with $(256, 128, 64, 32, 16)$ channels with increasing resolution from the bottleneck. Standard skip connections between the encoder output and the decoder output are used. The final output has one channel, which is passed through the sigmoid function to get the IMO probability. We trained our network when a small validation set loss curve flattens. We do not apply special gradient clipping or decay techniques. We used a learning rate of 2e-4 with an ADAM optimizer. The batch size of our training experiments is 32. On an Nvidia RTX 3090 GPU, the training speed is about 1 iteration per second. For the supervised baseline CNN, the network is trained in the exact setting. The only difference is that the ground truth IMO masks are given and the network can train longer because the ground truth masks can force the network to learn sharp boundaries as training progresses.

\fi
\end{document}